\documentclass[letterpaper, 10 pt, conference]{ieeeconf}
\IEEEoverridecommandlockouts    
\usepackage{graphicx,subcaption}
\usepackage{amsmath,amssymb} 
\usepackage{color}
\usepackage{times}
\usepackage{epsfig}
\let\labelindent\relax
\usepackage[inline]{enumitem}
\usepackage{makecell}  
\usepackage{multirow}
\usepackage{gensymb}
\usepackage{soul}
\usepackage[dvipsnames]{xcolor}
\usepackage{calc}
\usepackage{booktabs}
\usepackage{float}

\usepackage[hidelinks]{hyperref}

\usepackage{graphics}           
\usepackage{times}              
\usepackage{amsmath}            
\usepackage{amssymb}            
\usepackage{graphicx}
\usepackage{algorithm}
\usepackage[noend]{algpseudocode}
\usepackage{booktabs}
\usepackage{color}
\definecolor{instructioncolor}{rgb}{.5,.5,.5}

\usepackage[font=small]{caption}

\def\secref#1{Sec.~\ref{#1}}
\def\figref#1{Fig.~\ref{#1}}

\def\eqref#1{Eq.~(\ref{#1})}

\makeatletter
\usepackage{xspace}
\DeclareRobustCommand\onedot{\futurelet\@let@token\@onedot}
\def\@onedot{\ifx\@let@token.\else.\null\fi\xspace}

\def\etal{{et al}\onedot}
\makeatother

\def\etalcite#1{\etal~\cite{#1}}

\usepackage{array}
\newcolumntype{L}[1]{>{\raggedright\let\newline\\\arraybackslash\hspace{0pt}}m{#1}}
\newcolumntype{C}[1]{>{\centering\let\newline\\\arraybackslash\hspace{0pt}}m{#1}}
\newcolumntype{R}[1]{>{\raggedleft\let\newline\\\arraybackslash\hspace{0pt}}m{#1}}

\newcommand\hlf[1]{\textbf{\textcolor{red}{#1}}} 
\newcommand\hls[1]{\textbf{\textcolor{blue}{#1}}}  

\makeatletter
\newcommand{\printfnsymbol}[1]{%
  \textsuperscript{\@fnsymbol{#1}}%
}
\makeatother

\let\oldenumerate\enumerate
\renewcommand{\enumerate}{
\oldenumerate
\setlength{\itemsep}{1.2pt}
\setlength{\parskip}{0pt}
\setlength{\parsep}{0pt}
}

\newenvironment{tight_itemize}{
\begin{itemize}
  \setlength{\itemsep}{0pt}
  \setlength{\parskip}{0pt}
}{\end{itemize}}

\title{\LARGE \bf Learning-Based Dimensionality Reduction \\ for Computing Compact and Effective Local Feature Descriptors}

\author{Hao Dong \and Xieyuanli Chen \and Mihai Dusmanu \and Viktor Larsson \and Marc Pollefeys \and Cyrill Stachniss
  \thanks{H. Dong, M. Dusmanu, and M. Pollefeys are with ETH Z\"urich, Switzerland. X. Chen and C. Stachniss are with the University of Bonn, Germany. V. Larsson is with Lund University, Sweden. M. Pollefeys is additionally with Microsoft. C. Stachniss is additionally with the Department of Engineering Science at the University of Oxford, UK, and with the Lamarr Institute for Machine Learning and Artificial Intelligence, Germany.
  }%
}

\begin{document}
\maketitle
\thispagestyle{empty}
\pagestyle{empty}



\begin{abstract}
A distinctive representation of image patches in form of features is a key component of many computer vision and robotics tasks, such as image matching, image retrieval, and visual localization. State-of-the-art descriptors, from hand-crafted descriptors such as SIFT to learned ones such as HardNet, are usually high dimensional; 128 dimensions or even more. The higher the dimensionality, the larger the memory consumption and computational time for approaches using such descriptors. In this paper, we investigate multi-layer perceptrons (MLPs) to extract low-dimensional but high-quality descriptors. We thoroughly analyze our method in unsupervised, self-supervised, and supervised settings, and evaluate the dimensionality reduction results on four representative descriptors. We consider different applications, including visual localization, patch verification, image matching and retrieval. The experiments show that our lightweight MLPs achieve better dimensionality reduction than PCA. The lower-dimensional descriptors generated by our approach outperform the original higher-dimensional descriptors in downstream tasks, especially for the hand-crafted ones. The code will be available at \href{https://github.com/PRBonn/descriptor-dr}{https://github.com/PRBonn/descriptor-dr}. 
\end{abstract}

\section{Introduction}
Local feature descriptors~\cite{sift2004,surf2006,orb2011} are used to represent characteristics of image patches and are designed to be robust to partial occlusions, viewpoint changes, and variations in illumination. They play an essential role in many robotics applications such as robot localization~\cite{7572201}, object recognition~\cite{1211479}, and image retrieval~\cite{937561}.

The traditional pipeline for local feature extraction often starts by detecting the position, scale, and orientation of keypoints in the image. 
Then, a normalized image patch is extracted with respect to the estimated keypoint, which usually provides the basis for the descriptor computation. Distinctive and invariant keypoints and descriptors are key to achieving good performance in the subsequent matching, retrieval, and localization tasks.
This paper focuses on the descriptor part of this pipeline, specifically the dimensionality reduction of local feature descriptors to generate compact and at the same time effective features.

\begin{figure}[tb!]
\begin{center}
\includegraphics[width=0.9\linewidth]{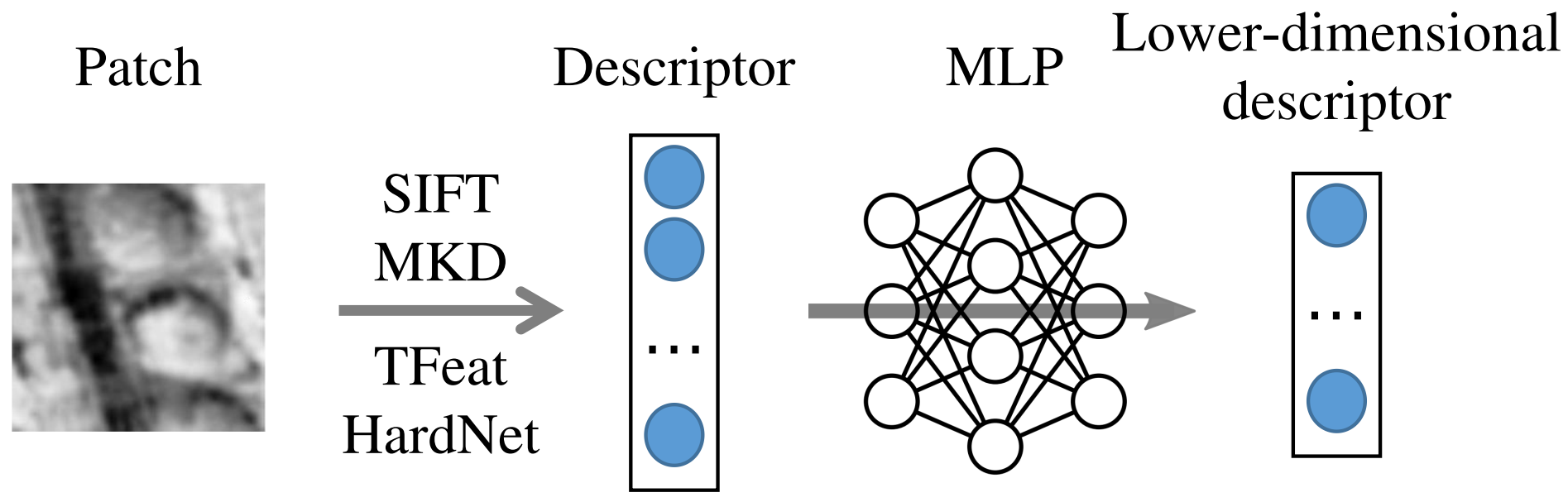}
\end{center}
\vspace{-3mm}
\caption{Overview of our approach. We first compute descriptors of given image patches. Then an MLP-based network is used for dimensionality reduction. We aim to learn an MLP-based projection better than PCA to generate lower-dimensional descriptors.}
\label{fig:triplet}
\vspace{-6mm}
\end{figure}

Multiple visual feature descriptors have been introduced in the literature.
Among all the hand-crafted descriptors, SIFT~\cite{sift2004} is one of the most famous because of its robustness under blurring, translation, rotation, and scale changes. 
With the advent of neural networks, more and more learning-based descriptors have been proposed~\cite{l2net2017,deepcompare2015,convexoptim2014,hardnet2017} and pushed the state-of-the-art forward in benchmarks for image matching, patch verification, and image retrieval. While the results are promising, a common issue for both hand-crafted and learned methods is that the dimensionality of the generated descriptors is usually high. When the image database increases, substantial time and space might be needed for computing and storing such high-dimensional descriptors. That may hinder the application to mobile and resource-constraint robots. Several principal component analysis (PCA)~\cite{Jolliffe2002Principal} based dimensionality reduction methods have been proposed to alleviate this problem. For example, Valenzuela~\etal~\cite{6782160} apply PCA to reduce the dimensionality of SIFT and SURF descriptors. Instead of the original SIFT’s smoothed weighted histograms, Ke~\etal~\cite{1315206} apply PCA to patches for generating lighter descriptors. PCA performs a linear projection from high- to low-dimensional descriptor space, which has limited capabilities of generating high-quality dimensionally reduced descriptors. Moreover, the components with low eigenvalues by PCA are not necessarily less important but down-weighted or even eliminated, which will cause information loss and performance degradation.

Unlike previous PCA-based methods, we aim to learn an MLP-based dimensionality reduction to better transform large descriptors into lighter ones. To this end, we thoroughly analyze the MLP-based network for dimensionality reduction and design three learning schemes, including unsupervised, self-supervised, and supervised methods. 
For the unsupervised scheme, we use an auto-encoder with reconstruction and distance losses to improve the projection.
While for the self-supervised one, we iteratively cluster descriptors using $k$-means and use cluster assignments as pseudo-labels to train the MLPs.
We also propose a supervised method that uses ground truth patch labels with triplet loss to supervise the MLPs generating more distinctive lower-dimensional descriptors. 
We evaluate our MLP-based method with all proposed learning schemes on four common descriptors: SIFT~\cite{sift2004}, MKD~\cite{Mukundan2018UnderstandingAI}, TFeat~\cite{tfeat2016}, and HardNet~\cite{hardnet2017}. 
We train our network only on one dataset and apply it directly to other datasets with different downstream tasks, including visual localization, patch verification, image matching, and image retrieval. The experimental results show that our method consistently outperforms the PCA-based method with a strong generalization ability.
Furthermore, using the lower-dimensional descriptors generated by our supervised MLP, we achieve even better performance in downstream tasks than the original higher-dimensional descriptors, especially for hand-crafted ones with faster speed and less memory consumption.
Overall, our contributions are as follows:
\begin{tight_itemize}
    \item We propose and evaluate an MLP-based network for descriptor dimensionality reduction and show its superiority over PCA on multiple descriptors in various tasks;
    \item We demonstrate that the lighter descriptors by our supervised MLP projection achieve even better performance in downstream tasks than the original descriptors;
    \item We thoroughly analyze the improvement of using our method for different descriptors on multiple datasets and show a good generalization of our method.
\end{tight_itemize}

\section{Related Work}
Various local image descriptors have been proposed over the past decades ranging from hand-crafted ones like SIFT~\cite{sift2004}, BRIEF~\cite{brief2012}, and ORB~\cite{orb2011} to learned ones like TFeat~\cite{tfeat2016}, HardNet~\cite{hardnet2017}, and SOSNet~\cite{sosnet2019cvpr}. Hand-crafted descriptors are typically based on human insights into which qualities are invariant under certain transformations, such as differential or moment invariants, correlations, and gradients histograms. For example, SIFT by Lowe~\cite{sift2004} generates descriptors based on the gradient distribution in the detected patches. BRIEF by Calonder~\etal~\cite{brief2012} uses simple binary intensity comparisons between pixels in an image patch. More details of classical hand-crafted descriptors can be found in surveys~\cite{ma2021ijcv,10.5555/2898913}. Benefiting from large-scale datasets~\cite{ubc2011}, learning-based descriptors recently achieved state-of-the-art performances~\cite{hardnet2017,tfeat2016,geodesc2018,Ebel19,7298948,deepdesc2015}. For example, TFeat by Balntas~\etal~\cite{tfeat2016} uses a CNN with hard-negative mining by anchor swap in the triplet loss to compute descriptors. In contrast, Tian~\etal~\cite{l2net2017} propose \mbox{L2-Net}, adding different error terms in the loss function to improve the distinctiveness of the descriptors. HardNet by Mishchuk~\etal~\cite{hardnet2017} also uses a simple triplet margin loss for hard negative mining which outperforms other descriptors with an advanced sampling procedure. Both, state-of-the-art hand-crafted and learning-based descriptors are often high dimensional. 

Dimensionality reduction techniques can be used to shorten the dimensionalities of feature descriptors. Traditional dimensionality reduction usually refers to reducing the dimension of data while keeping as much information as possible. Classical examples are backward elimination~\cite{mao2004orthogonal}, forward selection~\cite{mao2004orthogonal}, and random forests~\cite{breiman2001random}. Another type is to find a combination of new features to describe the data. For example. linear dimensionality reduction methods include PCA~\cite{Jolliffe2002Principal}, factor analysis~\cite{doi:10.1177/0095798418771807}, and linear discriminant analysis~\cite{lda}, and non-linear methods including Kernel PCA~\cite{10.1007/BFb0020217}, t-distributed stochastic neighbor embedding (t-SNE)~\cite{vanDerMaaten2008}, and isometric mapping~\cite{RASSIAS1999108}.
Among them, PCA~\cite{Jolliffe2002Principal} has been widely used for dimensionality reduction of image patch-based descriptors. For example, Gil~\etal~\cite{gil2006iros} and Valenzuela~\etal~\cite{6782160} use PCA to reduce the dimensionality of SIFT and SURF descriptors. Ke~\etal~\cite{1315206} apply PCA to the patches instead of using smoothed weighted histograms in SIFT. There are few works using neural networks for dimensionality reduction. The work most related to ours is the one by Loquercio~\etal~\cite{7989359}, which uses a supervised method to train a linear projection. However, they work on hand-crafted descriptors like FREAK~\cite{6247715} and focus only on the visual localization task. Different from them, we thoroughly analyze MLP-based non-linear projections using unsupervised, self-supervised, and supervised training schemes on multiple downstream tasks. Moreover, our method works on both hand-crafted descriptors such as SIFT~\cite{sift2004} and MKD~\cite{Mukundan2018UnderstandingAI}, and learned ones such as TFeat~\cite{tfeat2016} and HardNet~\cite{hardnet2017}, and shows a strong generalization ability.

\section{Methodology}
We aim to reduce the dimensionality of local feature descriptors using an MLP network. To understand the ability of the MLP-based method in this task, we investigate three learning schemes, including unsupervised, self-supervised, and supervised methods. This section will introduce the principle of each approach in detail.

\begin{figure}[t]
\begin{center}
\includegraphics[width=0.9\linewidth]{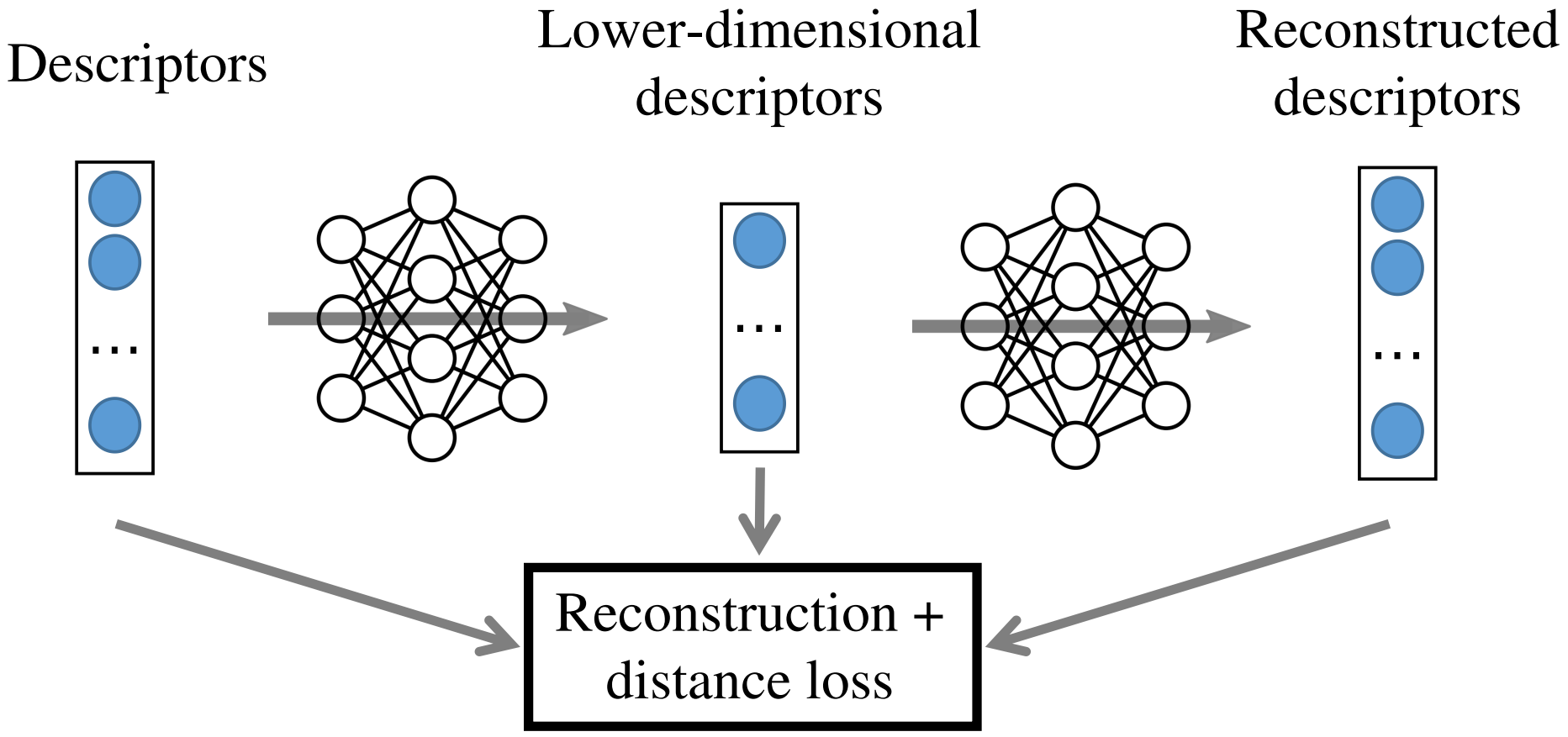}
\end{center}
\caption{The pipeline of using an auto-encoder for unsupervised dimensionality reduction. It consists of an encoder and a decoder. The encoder maps the original descriptors to lower-dimensional descriptors. The decoder tries to reconstruct the original descriptors from the projected lower-dimensional descriptors.}
\label{fig:ae}
\vspace{-5mm}
\end{figure}

\subsection{Unsupervised Reduction}
\label{sec:ae}
Auto-encoders~\cite{pmlr-v27-baldi12a} are unsupervised learning techniques for dimensionality reduction. Here, we use it to see whether an MLP-based network can learn a good projection from high to low dimensionality in an unsupervised way. To achieve fast and lightweight dimensionality reduction, we build our auto-encoder using MLPs with no more than two hidden layers as shown in~\figref{fig:ae}. Our auto-encoder consists of a symmetric encoder and decoder. The encoder projects the input feature into lower-dimensional embeddings, while the decoder reconstructs the original input from the lower-dimensional embeddings. By checking the consistency between inputs and outputs, the auto-encoder learns how to extract lower-dimensional descriptors from the original ones without labels. We apply the consistency constraints by minimizing the reconstruction loss between inputs and outputs.
The reconstruction loss $\mathcal{L}^R$ measures the differences between the input descriptors $\{\boldsymbol{x}_i\}_{i = 1}^N$ in a training mini-batch and the reconstructed output descriptors $\{\boldsymbol{x}_{i}^{\prime}\}_{i = 1}^N$ as

\begin{equation}
\mathcal{L}^R=\frac{1}{N}\sum_{i=1}^{N}\left\|\boldsymbol{x}_{i}-\boldsymbol{x}_{i}^{\prime}\right\|_2 .
\end{equation}

We propose an additional distance loss $\mathcal{L}^D$ for HardNet.
The distance loss calculates the difference between the distance of the original high-dimensional descriptors and the distance of the lower-dimensional descriptors in the embedding space. Given two descriptors $\boldsymbol{x}_{i}$ and $\boldsymbol{x}_{j}$, their corresponding lower-dimensional descriptors in the embedding space are $\boldsymbol{\hat{x}}_{i}$ and $\boldsymbol{\hat{x}}_{j}$. The loss $\mathcal{L}^D$ is to make the $\ell_2$ distance between the lower-dimensional descriptors $d\left(\boldsymbol{\hat{x}}_{i}, \boldsymbol{\hat{x}}_{j}\right)$ as similar as possible compared to that between the original descriptors $d\left(\boldsymbol{x}_{i}, \boldsymbol{x}_{j}\right)$
\begin{equation}
\mathcal{L}^D=\frac{1}{N(N-1)}\sqrt{\sum_{i=1}^{N}\sum_{j \neq i}\left(d\left(\boldsymbol{x}_{i}, \boldsymbol{x}_{j}\right)-d\left(\boldsymbol{\hat{x}}_{i}, \boldsymbol{\hat{x}}_{j}\right)\right)^{2}},
\end{equation}
where $d\left(\boldsymbol{x}_{i}, \boldsymbol{x}_{j}\right) = \|\boldsymbol{x}_{i}-\boldsymbol{x}_{j}\|_2$ is the $\ell_2$ distance and the loss is calculated on all different pairs in a mini-batch. The final loss for HardNet is a weighted sum of the reconstruction and distance losses	
\begin{equation}
\mathcal{L} = \mathcal{L}^R + \alpha \mathcal{L}^D,
\end{equation}
more explanation about this design is given in Sec.~\ref{sec:discuss}.

Note that we need no other information but image patches to train our auto-encoder. 
Moreover, the non-linearity of the auto-encoder allows it to learn better projection than PCA, thus generating better lower-dimensional descriptors. This will be shown in the experimental evaluation.

\subsection{Self-Supervised Reduction}
\label{sec:ss}

\begin{figure}[t]
\begin{center}
\includegraphics[width=\linewidth]{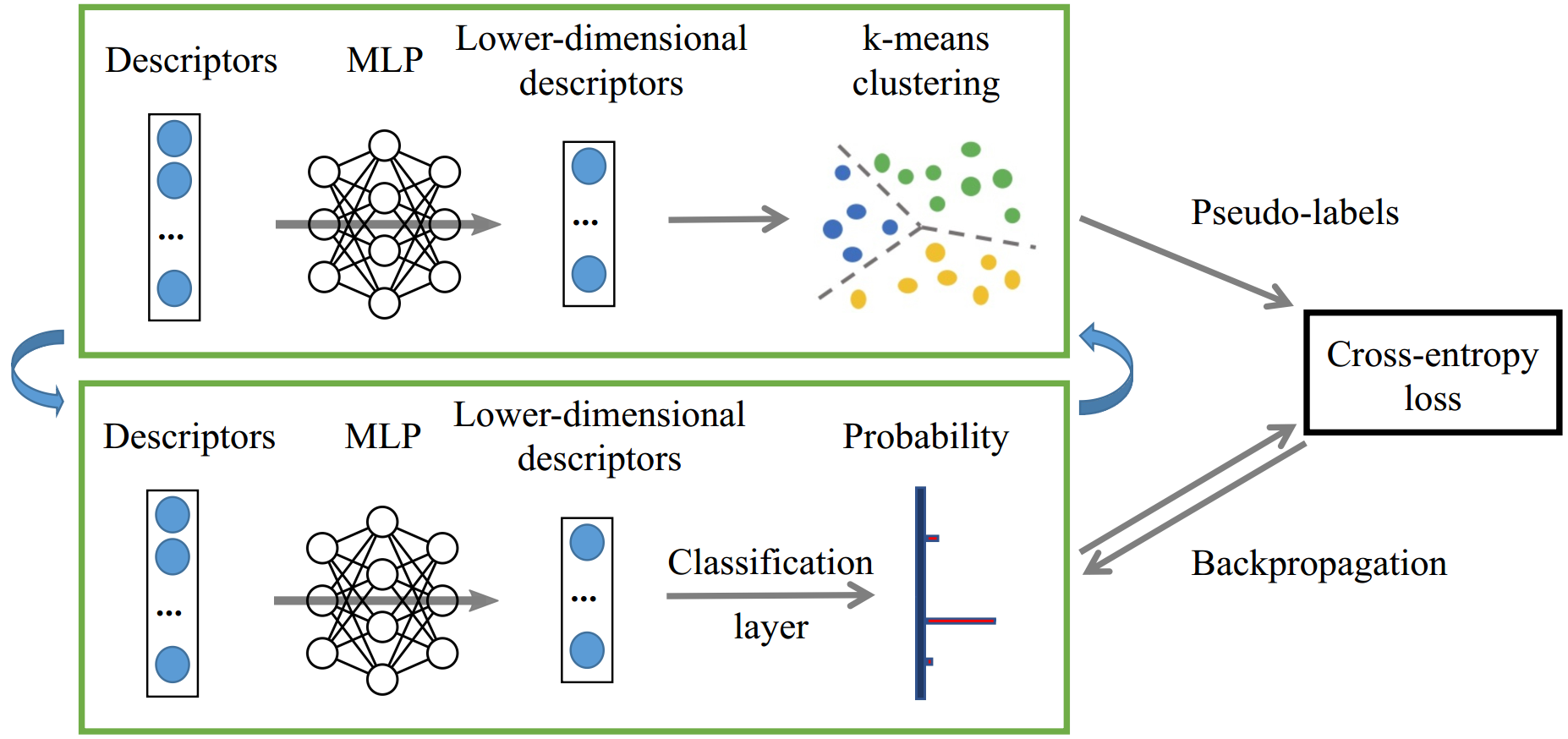}
\end{center}
\caption{Pipeline of our self-supervised method. We iteratively cluster descriptors and use the clustering assignments as pseudo-labels to train the network.}
\label{fig:ss}
\vspace{-2mm}
\end{figure}

Unlike unsupervised methods, self-supervised methods usually use traditional heuristic-based methods to generate pseudo-labels and guide networks to learn certain tasks. It combines human priors with learning-based methods to achieve good performance.
Inspired by deep feature clustering~\cite{caron2018deep,larsson2019fgsn}, we propose a self-supervised method for MLP-based dimensionality reduction. 
The main idea is to apply $k$-means clustering with descriptors and use the clustering assignments as pseudo-labels, i.e., descriptors in the same cluster considered to have the same label. 
A classification layer~\cite{8953658} is added during training to guide our MLPs learning to generate similar lower-dimensional descriptors if their high-dimensional ones have the same pseudo-label. In the first training epoch, we cluster the original descriptors into different groups and use them as labels to supervise our MLPs. From that on, we use the clusters of the lower-dimensional descriptors from the previous epoch as supervision. 
Given a set of descriptors extracted from image patches, $k$-means clustering generates a centroid matrix $\boldsymbol{C}$ and the clustering assignments $\boldsymbol{y}_{i}$ for each descriptor $\boldsymbol{x}_{i}$ by solving
\begin{equation}
\small{
\min _{C \in \mathbb{R}^{d \times k}} \frac{1}{N} \sum_{i=1}^{N} \min _{\boldsymbol{y}_{i} \in\{0,1\}^{k}}\left\|\boldsymbol{x}_{i}-\boldsymbol{C} \boldsymbol{y}_{i}\right\|_{2}^{2}~,~\textrm{s.t.}~\boldsymbol{y}_{i}^{\top} \boldsymbol{1}_{k}=1 ,
}
\end{equation}
where $\boldsymbol{1}_{k}$ is a vector whose all elements are $1$ and $\boldsymbol{y}_{i}$ is a one-hot vector.

The training loss used for our self-supervised method is the standard cross-entropy loss with the pseudo-labels as targets.

\subsection{Supervised Reduction}
\label{sec:triplet}

\begin{figure}[t]
\begin{center}
\includegraphics[width=\linewidth]{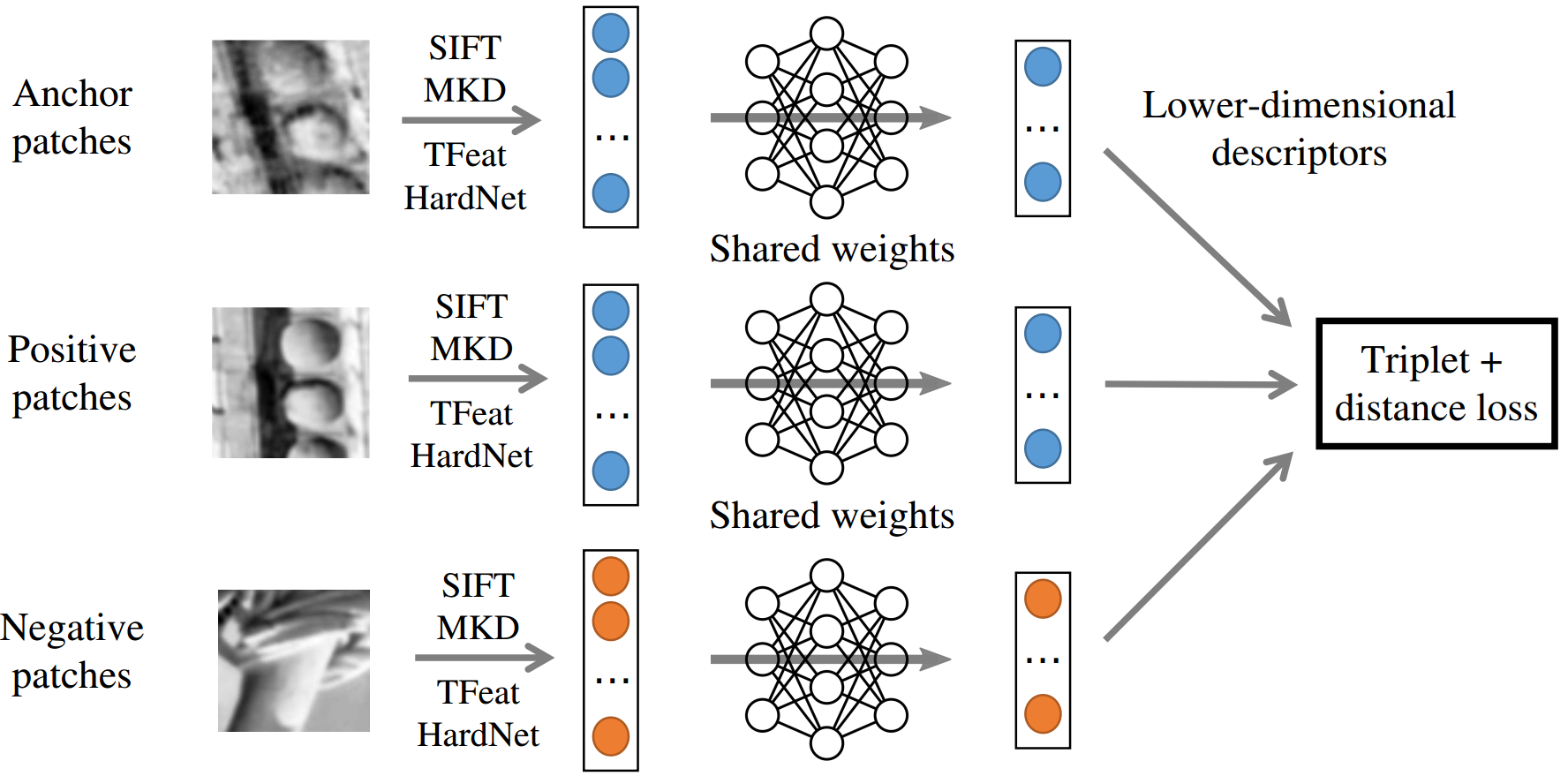}
\end{center}
\caption{Pipeline of the supervised method. A training sample includes an anchor with positive and negative patches. The extracted descriptors are fed into MLPs to get lower-dimensional descriptors and a triplet loss is applied to supervise the MLPs.}
\label{fig:triplet}
\vspace{-2mm}
\end{figure}

We next introduce our supervised method for dimensionality reduction.
As shown in~\figref{fig:triplet}, we use ground truth patch labels together with the triplet loss to train the MLP. A batch of matching patches is denoted as ${\{ {{{a}_i},{p}_i} \}}_{{i = 1 \dots N}}$, where $a$ stands for the anchor patch and $p$ for the positive patch. The non-matching negative patches ${\{ {{{n}_i}} \}}_{{i = 1 \dots N}}$ are sampled by the hardest-within-batch strategy as introduced by Mishchuk~\etalcite{hardnet2017}. For all training patches, we first use existing methods to generate higher-dimensional descriptors and feed them to our MLPs to generate low-dimensional descriptors where the triplet margin loss is applied
\begin{equation}
\small{
\mathcal{L}^T = \frac{1}{N}\sum_{i=1}^{N}{\max{(0, m + d(f(a_i), f(p_i)) - d(f(a_i), f(n_i)))}},
}
\end{equation}
where $f$ is the MLP-based projection. The main idea is to learn an MLP-based projection $f$ such that the distance between the anchor and positive descriptors $d(f(a_i), f(p_i))$ is smaller than that between the anchor and negative descriptors $d(f(a_i), f(n_i))$ with a margin $m$ in the lower-dimensional embedding space.
We also add a distance loss term for HardNet as used in the auto-encoder to target for the similarity of the distance between input descriptors and embedding descriptors in a batch. The final loss for HardNet is a weighted sum of the triplet margin and distance losses
\begin{equation}
\mathcal{L} = \mathcal{L}^T + \beta \mathcal{L}^D.
\end{equation}

\subsection{Training and Parameters}
\label{sec:imple}
We use the UBC Phototour Liberty dataset~\cite{ubc2011} for training. After training our networks on this dataset, we apply the trained model to other tasks and datasets without fine-tuning.

For the auto-encoder, we train the network for $5$ epochs using Adam~\cite{adam2014} with a learning rate of $0.001$ and a batch size of $1024$. We choose the distance loss weighting factor \mbox{$\alpha = 0.1$} for HardNet.
For the self-supervised method, we train the network for $200$ epochs using Adam with a learning rate of $0.001$ and a batch size of $256$. The number of cluster for $k$-means is $100\,000$. The clustering is repeated and the classification layer is re-initialized every $10$ epochs. 
For the supervised method, we choose the margin $m=1$ and train the network for $10$ epochs using Adam with a learning rate of $0.001$ and a batch size of $1024$. The learning rate is linearly decayed to zero within $10$ epochs. We choose the weighting of the distance loss $\beta = 3$ for HardNet. 
We use ReLU followed by batch normalization~\cite{batch2015} after each linear layer except the last one. The embeddings are $\ell_2$-normalized. 
For hand-crafted descriptors, we use two hidden layers for our MLPs, while for learning-based ones we use only one hidden layer. 
More detailed parameters and network architectures for each method can be found in our open-source implementation.
For PCA as the baseline, we use the implementation from scikit-learn~\cite{scikit-learn}. 

\section{Experiments}
We present our experiments to show the capabilities of our MLP-based methods for different tasks. We choose SIFT~\cite{sift2004}, MKD~\cite{Mukundan2018UnderstandingAI}, TFeat~\cite{tfeat2016}, and HardNet~\cite{hardnet2017} as the base descriptors. Their original dimensions are all $128$. We convert these descriptors to lower dimensions of $64$, $32$, $24$, and $16$ and apply them on three publicly available datasets, HPatches~\cite{hpatches2017}, Aachen Day-Night v1.1~\cite{9229078}, and InLoc~\cite{Taira2018CVPR} for different downstream tasks, including visual localization, patch verification, image matching, and patch retrieval. We name our unsupervised method `Ours-US', the self-supervised method `Ours-SS', and the supervised method `Ours-SV' in all experiments.

\begin{table}[t]
\begin{center}
\scriptsize{
\setlength{\tabcolsep}{2pt}
\begin{tabular}{l|c|c|c|c|c|c|c|c}& \multicolumn{2}{c|}{\textbf{Aachen Day-Night v1.1}}&  \multicolumn{2}{c|}{\textbf{InLoc}}\\

& {day}& {night}& {duc1}& {duc2}\\ \cline{2-5}

\multicolumn{1}{r|}{\begin{tabular}[c]{@{}r@{}}distance [m]\\ orient. [deg]\end{tabular}} & {\begin{tabular}[c]{@{}c@{}}0.25\,/\,0.5\,/\,5.0\\ 2\,/\,5\,/\,10\end{tabular}} & {\begin{tabular}[c]{@{}c@{}}0.5\,/\,1.0\,/\,5.0\\ 2\,/\,5\,/\,10\end{tabular}} & {\begin{tabular}[c]{@{}c@{}}0.25\,/\,0.5\,/\,1.0\\ N\,/\,A\end{tabular}} &{\begin{tabular}[c]{@{}c@{}}0.25\,/\,0.5\,/\,1.0\\N\,/\,A\end{tabular}}\\ \hline

{SIFT} & 88.1\,/\,94.7\,/\,98.4  &	64.9\,/\,77.5\,/\,92.1 &  31.3\,/\,\hlf{46.0}\,/\,56.1 & 21.4\,/\,33.6\,/\,43.5 \\ \hline

{PCA-16} & 84.5\,/\,90.7\,/\,95.9 & \hls{38.2}\,/\,\hls{47.1}\,/\,58.6 & 19.7\,/\,32.3\,/\,37.9 & 11.5\,/\,20.6\,/\,25.2\\ 

{Ours-US-16} & 84.5\,/\,\hls{92.4}\,/\,96.0 & 35.6\,/\,45.0\,/\,\hls{60.7} & 22.2\,/\,33.8\,/\,39.4 & 13.0\,/\,20.6\,/\,24.4\\

{Ours-SS-16} & 85.1\,/\,90.9\,/\,95.6 & 34.0\,/\,41.9\,/\,57.6 & 23.7\,/\,35.4\,/\,41.4 & \hls{16.0}\,/\,22.1\,/\,26.0\\ 

{Ours-SV-16} & \hls{85.9}\,/\,91.5\,/\,\hls{96.2} & 34.6\,/\,46.1\,/\,58.6 & \hls{24.2}\,/\,\hls{36.9}\,/\,\hls{43.9} & 15.3\,/\,\hls{26.7}\,/\,\hls{30.5}\\ \hline

{PCA-24} & 86.2\,/\,93.2\,/\,\hls{97.5} & 50.3\,/\,60.7\,/\,77.5 & 24.2\,/\,34.8\,/\,46.5 & 15.3\,/\,26.0\,/\,29.8\\

{Ours-US-24} & 86.8\,/\,93.3\,/\,97.2 & 50.3\,/\,\hls{64.4}\,/\,\hls{78.5} & 24.2\,/\,39.9\,/\,48.5 & \hls{21.4}\,/\,29.8\,/\,35.9\\ 

{Ours-SS-24} & 87.7\,/\,93.4\,/\,97.3 & 48.7\,/\,59.7\,/\,74.9 & 25.8\,/\,42.4\,/\,48.5 & 17.6\,/\,26.7\,/\,32.8\\ 

{Ours-SV-24} & \hls{87.9}\,/\,\hls{93.8}\,/\,97.3 & \hls{51.3}\,/\,62.8\,/\,\hls{78.5} & \hls{30.8}\,/\,\hls{43.4}\,/\,\hls{52.0} & \hls{21.4}\,/\,\hls{34.4}\,/\,\hls{41.2}\\ \hline

{PCA-32} & 87.1\,/\,93.0\,/\,97.6 & 56.5\,/\,69.1\,/\,80.6 & 28.3\,/\,40.9\,/\,53.5 & 21.4\,/\,29.0\,/\,35.9\\ 

{Ours-US-32} & 87.0\,/\,94.1\,/\,97.8 & \hls{58.1}\,/\,71.2\,/\,\hls{83.2} & 25.8\,/\,40.9\,/\,52.5 & 19.8\,/\,32.1\,/\,37.4\\

{Ours-SS-32} & 87.9\,/\,93.8\,/\,98.1 & 57.1\,/\,66.0\,/\,82.7 & \hls{30.8}\,/\,\hls{43.9}\,/\,\hls{55.6} & 24.4\,/\,32.8\,/\,38.2\\

{Ours-SV-32} & \hls{88.7}\,/\,\hls{94.4}\,/\,\hls{98.3} & \hls{58.1}\,/\,\hls{71.7}\,/\,82.7 & 29.8\,/\,42.4\,/\,54.5 & \hlf{27.5}\,/\,\hls{38.9}\,/\,\hls{44.3}\\ \hline

{PCA-64} & 87.1\,/\,94.3\,/\,98.2 & 60.2\,/\,74.3\,/\,88.5 & 28.8\,/\,41.4\,/\,53.0 & 19.8\,/\,31.3\,/\,40.5\\ 

{Ours-US-64} & 87.6\,/\,\hlf{95.3}\,/\,98.5 & \hlf{66.0}\,/\,\hlf{79.6}\,/\,90.6 & 31.3\,/\,43.9\,/\,57.1 & \hls{26.7}\,/\,37.4\,/\,45.0\\ 

{Ours-SS-64} & 87.9\,/\,94.9\,/\,98.2 & 60.2\,/\,75.9\,/\,89.0 & 30.8\,/\,43.4\,/\,54.5 & 23.7\,/\,\hlf{39.7}\,/\,\hlf{45.8}\\ 

{Ours-SV-64} & \hlf{89.1}\,/\,94.8\,/\,\hlf{98.8} & 63.4\,/\,\hlf{79.6}\,/\,\hlf{92.7} & \hlf{33.3}\,/\,\hlf{46.0}\,/\,\hlf{57.6} & 26.0\,/\,\hlf{39.7}\,/\,\hlf{45.8}\\ \hline



\end{tabular}

}
\end{center}
\vspace{-3mm}
\caption{Evaluation of the localization on the Aachen Day-Night v1.1 and InLoc datasets with SIFT~\cite{sift2004} features. We report the recall~[\%] at different distances and orientation thresholds. The overall best results are in \hlf{red} and the best results with the same low dimension are in \hls{blue}.
}
\vspace{-2mm}
\label{tab:loc_results:comparison}
\end{table}
\begin{table}[t]
\begin{center}
\scriptsize{
\setlength{\tabcolsep}{2pt}
\begin{tabular}{l|c|c|c|c|c|c|c|c}& \multicolumn{2}{c|}{\textbf{Aachen Day-Night v1.1}}&  \multicolumn{2}{c|}{\textbf{InLoc}}\\

& {day}& {night}& {duc1}& {duc2}\\ \cline{2-5}

\multicolumn{1}{r|}{\begin{tabular}[c]{@{}r@{}}distance [m]\\ orient. [deg]\end{tabular}} & {\begin{tabular}[c]{@{}c@{}}0.25\,/\,0.5\,/\,5.0\\ 2\,/\,5\,/\,10\end{tabular}} & {\begin{tabular}[c]{@{}c@{}}0.5\,/\,1.0\,/\,5.0\\ 2\,/\,5\,/\,10\end{tabular}} & {\begin{tabular}[c]{@{}c@{}}0.25\,/\,0.5\,/\,1.0\\N\,/\,A\end{tabular}} &{\begin{tabular}[c]{@{}c@{}}0.25\,/\,0.5\,/\,1.0\\N\,/\,A\end{tabular}}\\ \hline

{MKD} & 88.7\,/\,95.1\,/\,98.8  & \hlf{67.0}\,/\,79.6\,/\,\hlf{92.7} &  \hlf{33.3}\,/\,\hlf{50.5}\,/\,\hlf{65.2} & 26.0\,/\,40.5\,/\,49.6 \\ \hline

{PCA-16} & \hls{85.7}\,/\,92.5\,/\,96.1 & 37.2\,/\,\hls{45.0}\,/\,\hls{56.0} & 23.7\,/\,35.4\,/\,42.9 & 17.6\,/\,25.2\,/\,28.2\\ 

{Ours-US-16} & 84.6\,/\,91.3\,/\,95.5 & 34.0\,/\,41.4\,/\,52.4 & 22.2\,/\,32.3\,/\,39.4 & 14.5\,/\,19.8\,/\,25.2\\

{Ours-SS-16} & 85.0\,/\,92.2\,/\,96.5 & 36.6\,/\,44.0\,/\,55.5 & \hls{24.2}\,/\,34.8\,/\,43.9 & 14.5\,/\,26.0\,/\,31.3\\ 

{Ours-SV-16} & \hls{85.7}\,/\,\hls{93.0}\,/\,\hls{96.8} & \hls{37.7}\,/\,42.9\,/\,53.9 & 23.2\,/\,\hls{35.9}\,/\,\hls{45.5} & \hls{21.4}\,/\,\hls{28.2}\,/\,\hls{32.8}\\ \hline

{PCA-24} & 87.4\,/\,93.8\,/\,97.5 & 49.2\,/\,59.2\,/\,70.7 & 26.8\,/\,37.9\,/\,48.0 & 20.6\,/\,31.3\,/\,37.4\\

{Ours-US-24} & 87.6\,/\,93.8\,/\,97.2 & 47.6\,/\,60.2\,/\,73.8 & 24.7\,/\,38.4\,/\,49.0 & 18.3\,/\,27.5\,/\,35.1\\ 

{Ours-SS-24} & \hls{88.0}\,/\,\hls{94.1}\,/\,\hls{97.9} & 51.3\,/\,\hls{64.4}\,/\,\hls{76.4} & 27.8\,/\,\hls{40.4}\,/\,51.5 & 18.3\,/\,30.5\,/\,38.2\\ 

{Ours-SV-24} & 87.5\,/\,93.7\,/\,97.3 & \hls{53.9}\,/\,63.9\,/\,75.4 & \hls{29.3}\,/\,\hls{40.4}\,/\,\hls{53.5} & \hls{22.9}\,/\,\hls{35.9}\,/\,\hls{39.7}\\ \hline

{PCA-32} & 87.6\,/\,93.9\,/\,98.2 & 52.4\,/\,67.0\,/\,81.7 & 30.8\,/\,42.4\,/\,51.5 & 26.0\,/\,36.6\,/\,43.5\\ 

{Ours-US-32} & 87.6\,/\,94.2\,/\,98.4 & 59.2\,/\,71.2\,/\,\hls{85.9} & 27.8\,/\,40.9\,/\,55.6 & 27.5\,/\,37.4\,/\,42.0\\

{Ours-SS-32} & \hls{88.6}\,/\,\hls{94.7}\,/\,98.2 & 56.0\,/\,70.7\,/\,83.2 & 30.3\,/\,\hls{44.4}\,/\,\hls{57.6} & \hls{28.2}\,/\,\hls{42.7}\,/\,\hls{47.3}\\

{Ours-SV-32} & 88.3\,/\,94.5\,/\,\hls{98.7} & \hls{60.7}\,/\,\hls{73.8}\,/\,\hls{85.9} & \hls{31.3}\,/\,43.4\,/\,56.6 & 26.7\,/\,38.9\,/\,46.6\\ \hline

{PCA-64} & 88.1\,/\,94.9\,/\,98.5 & 62.3\,/\,\hlf{80.6}\,/\,92.1 & \hls{31.3}\,/\,47.5\,/\,61.1 & 26.0\,/\,41.2\,/\,48.1\\ 

{Ours-US-64} & 88.6\,/\,94.5\,/\,98.5 & \hls{65.4}\,/\,\hlf{80.6}\,/\,\hlf{92.7} & 30.8\,/\,47.5\,/\,59.6 & 28.2\,/\,42.0\,/\,48.1\\ 

{Ours-SS-64} & 88.8\,/\,\hlf{95.6}\,/\,\hlf{98.9} & 62.3\,/\,79.6\,/\,92.1 & \hls{31.3}\,/\,44.9\,/\,58.1 & \hlf{30.5}\,/\,44.3\,/\,50.4\\ 

{Ours-SV-64} & \hlf{89.0}\,/\,95.1\,/\,98.5 & 63.4\,/\,79.6\,/\,90.1 & \hls{31.3}\,/\,\hls{49.5}\,/\,\hls{61.6} & 28.2\,/\,\hlf{45.0}\,/\,\hlf{51.9}\\ \hline



\end{tabular}

}
\end{center}
\vspace{-3mm}
\caption{Evaluation of the localization on the Aachen Day-Night v1.1 and InLoc datasets with MKD~\cite{Mukundan2018UnderstandingAI} features. 
}
\vspace{-5mm}
\label{tab:loc_results:comparison_mkd}
\end{table}

\begin{table}[t]
\begin{center}
\scriptsize{
\setlength{\tabcolsep}{2pt}
\begin{tabular}{l|c|c|c|c|c|c|c|c}& \multicolumn{2}{c|}{\textbf{Aachen Day-Night v1.1}}&  \multicolumn{2}{c|}{\textbf{InLoc}}\\

& {day}& {night}& {duc1}& {duc2}\\ \cline{2-5}

\multicolumn{1}{r|}{\begin{tabular}[c]{@{}r@{}}distance [m]\\ orient. [deg]\end{tabular}} & {\begin{tabular}[c]{@{}c@{}}0.25\,/\,0.5\,/\,5.0\\ 2\,/\,5\,/\,10\end{tabular}} & {\begin{tabular}[c]{@{}c@{}}0.5\,/\,1.0\,/\,5.0\\ 2\,/\,5\,/\,10\end{tabular}} & {\begin{tabular}[c]{@{}c@{}}0.25\,/\,0.5\,/\,1.0\\N\,/\,A\end{tabular}} &{\begin{tabular}[c]{@{}c@{}}0.25\,/\,0.5\,/\,1.0\\N\,/\,A\end{tabular}}\\ \hline

{TFeat} & 87.4\,/\,93.9\,/\,98.2 & 53.4\,/\,72.8\,/\,83.8 &  32.8\,/\,\hlf{44.9}\,/\,55.6 &	24.4\,/\,41.2\,/\,45.8 \\ \hline

{PCA-16} & 82.6\,/\,89.4\,/\,94.7 & 30.4\,/\,37.7\,/\,\hls{49.7} & 20.7\,/\,29.8\,/\,39.9 &	\hls{16.0}\,/\,\hls{24.4}\,/\,\hls{29.8}\\ 

{Ours-US-16} & \hls{84.5}\,/\,\hls{91.1}\,/\,95.5 & 30.4\,/\,\hls{39.8}\,/\,\hls{49.7} & 21.2\,/\,31.3\,/\,39.9 &	13.7\,/\,19.1\,/\,24.4\\

{Ours-SS-16} & 84.0\,/\,89.9\,/\,95.4 & \hls{31.4}\,/\,36.6\,/\,47.1 & 21.2\,/\,30.3\,/\,37.9 &	10.7\,/\,19.8\,/\,22.1\\ 

{Ours-SV-16} & 83.9\,/\,90.8\,/\,\hls{95.6} & 28.8\,/\,34.0\,/\,47.1 & 	\hls{21.7}\,/\,\hls{32.3}\,/\,\hls{41.9} &	14.5\,/\,20.6\,/\,24.4\\ \hline

{PCA-24} & 86.3\,/\,93.2\,/\,96.6 & \hls{42.9}\,/\,55.0\,/\,67.0 & 23.7\,/\,38.9\,/\,49.0 &	20.6\,/\,30.5\,/\,35.1\\

{Ours-US-24} & 86.3\,/\,\hls{93.8}\,/\,\hls{97.1} & 42.4\,/\,55.0\,/\,63.4 & 26.3\,/\,37.9\,/\,48.0 &	19.1\,/\,31.3\,/\,35.9\\ 

{Ours-SS-24} & \hls{86.8}\,/\,92.7\,/\,96.4 & 36.6\,/\,46.6\,/\,\hls{67.5} & 27.3\,/\,38.9\,/\,48.0 &	\hls{22.9}\,/\,\hls{32.1}\,/\,\hls{36.6}\\ 

{Ours-SV-24} & 86.0\,/\,92.5\,/\,96.8 & \hls{42.9}\,/\,\hls{56.0}\,/\,\hls{67.5} & \hls{27.8}\,/\,\hls{39.4}\,/\,\hls{50.5} &	19.1\,/\,28.2\,/\,34.4\\ \hline

{PCA-32} & 87.3\,/\,93.7\,/\,97.2 & 47.1\,/\,61.8\,/\,71.7 & 25.8\,/\,38.9\,/\,52.0 &	22.9\,/\,33.6\,/\,\hls{42.7}\\ 

{Ours-US-32} & 87.6\,/\,\hls{94.3}\,/\,\hls{98.1} & \hls{48.2}\,/\,61.3\,/\,\hls{75.4} & \hls{30.3}\,/\,40.9\,/\,\hls{53.0} &	22.1\,/\,\hls{35.9}\,/\,40.5\\

{Ours-SS-32} & 87.3\,/\,93.1\,/\,97.7 & 45.5\,/\,59.2\,/\,73.3 & \hls{30.3}\,/\,40.4\,/\,50.5 &	22.9\,/\,35.1\,/\,41.2\\

{Ours-SV-32} & \hls{88.1}\,/\,94.2\,/\,\hls{98.1} & \hls{48.2}\,/\,\hls{62.3}\,/\,72.8 & \hls{30.3}\,/\,\hls{41.4}\,/\,51.5 &	\hls{23.7}\,/\,35.1\,/\,39.7\\ \hline

{PCA-64} & 88.2\,/\,\hlf{94.4}\,/\,98.2 & 52.9\,/\,70.2\,/\,79.6 & 29.3\,/\,43.4\,/\,54.0 &	28.2\,/\,38.9\,/\,46.6\\ 

{Ours-US-64} & 87.6\,/\,93.9\,/\,98.3 & 55.0\,/\,\hlf{73.3}\,/\,\hlf{84.8} & 29.3\,/\,41.9\,/\,58.6 &	26.0\,/\,39.7\,/\,\hlf{50.4}\\ 

{Ours-SS-64} & \hlf{89.0}\,/\,93.9\,/\,\hlf{98.7} & \hlf{56.5}\,/\,69.1\,/\,82.7 & 	\hlf{33.3}\,/\,\hls{44.4}\,/\,\hlf{60.1} &	23.7\,/\,39.7\,/\,46.6\\ 

{Ours-SV-64} & 87.0\,/\,\hlf{94.4}\,/\,98.4 & 55.0\,/\,71.2\,/\,\hlf{84.8} & 32.3\,/\,\hls{44.4}\,/\,57.1 &	\hlf{32.1}\,/\,\hlf{44.3}\,/\,\hlf{50.4}\\ \hline



\end{tabular}

}
\end{center}
\vspace{-3mm}
\caption{Evaluation of the localization on the Aachen Day-Night v1.1 and InLoc datasets with TFeat~\cite{tfeat2016} features. 
}
\vspace{-2mm}
\label{tab:loc_results:comparison_tfeat}
\end{table}
\begin{table}[t]
\begin{center}
\scriptsize{
\setlength{\tabcolsep}{2pt}

\begin{tabular}{l|c|c|c|c|c|c|c|c}& \multicolumn{2}{c|}{\textbf{Aachen Day-Night v1.1}}&  \multicolumn{2}{c|}{\textbf{InLoc}}\\

& {day}& {night}& {duc1}& {duc2}\\ \cline{2-5}

\multicolumn{1}{r|}{\begin{tabular}[c]{@{}r@{}}distance [m]\\ orient. [deg]\end{tabular}} & {\begin{tabular}[c]{@{}c@{}}0.25\,/\,0.5\,/\,5.0\\ 2\,/\,5\,/\,10\end{tabular}} & {\begin{tabular}[c]{@{}c@{}}0.5\,/\,1.0\,/\,5.0\\ 2\,/\,5\,/\,10\end{tabular}} & {\begin{tabular}[c]{@{}c@{}}0.25\,/\,0.5\,/\,1.0\\N\,/\,A\end{tabular}} &{\begin{tabular}[c]{@{}c@{}}0.25\,/\,0.5\,/\,1.0\\N\,/\,A\end{tabular}}\\ \hline

{HardNet} & 88.8\,/\,95.6\,/\,\hlf{99.3}  & \hlf{63.9}\,/\,\hlf{81.2}\,/\,92.7 &  37.9\,/\,56.6\,/\,70.7 & 31.3\,/\,44.3\,/\,53.4 \\ \hline

{PCA-16} & \hls{84.7}\,/\,\hls{91.6}\,/\,\hls{96.2} & 27.2\,/\,34.6\,/\,46.1 & 21.7\,/\,\hls{36.4}\,/\,\hls{43.9} & \hls{15.3}\,/\,\hls{24.4}\,/\,\hls{29.0}\\ 

{Ours-US-16} & 84.1\,/\,91.1\,/\,95.0 & 29.8\,/\,34.0\,/\,45.0 & 20.7\,/\,33.3\,/\,40.4 & 12.2\,/\,19.8\,/\,24.4\\

{Ours-SS-16} & 84.0\,/\,90.7\,/\,96.0 & 28.3\,/\,35.1\,/\,42.4 & 22.7\,/\,32.8\,/\,41.9 & 12.2\,/\,19.8\,/\,22.9\\ 

{Ours-SV-16} & 	84.0\,/\,91.3\,/\,95.8 & \hls{30.4}\,/\,\hls{36.1}\,/\,\hls{47.6} & \hls{24.7}\,/\,34.8\,/\,42.4 & 13.0\,/\,22.9\,/\,26.7\\ \hline

{PCA-24} & 86.9\,/\,93.4\,/\,\hls{98.3} & 44.5\,/\,55.0\,/\,69.6 & 29.3\,/\,\hls{42.9}\,/\,53.0 & 22.1\,/\,\hls{35.9}\,/\,\hls{38.9}\\

{Ours-US-24} & \hls{87.4}\,/\,\hls{94.1}\,/\,97.2 & 40.8\,/\,\hls{58.6}\,/\,69.6 & 28.3\,/\,38.9\,/\,52.5 & 18.3\,/\,29.0\,/\,35.9\\ 

{Ours-SS-24} & 86.4\,/\,93.6\,/\,97.6 & 46.1\,/\,57.1\,/\,66.5 & \hls{29.8}\,/\,41.4\,/\,50.0 & \hls{22.9}\,/\,32.8\,/\,35.9\\ 

{Ours-SV-24} & \hls{87.4}\,/\,\hls{94.1}\,/\,97.8 & \hls{48.7}\,/\,\hls{58.6}\,/\,\hls{71.2} & 25.8\,/\,\hls{42.9}\,/\,\hls{55.6} & 21.4\,/\,31.3\,/\,36.6\\ \hline

{PCA-32} & 88.0\,/\,94.8\,/\,\hls{98.9} & \hls{55.0}\,/\,\hls{69.6}\,/\,82.7 & 32.3\,/\,49.0\,/\,60.1 & \hls{28.2}\,/\,38.9\,/\,44.3\\ 

{Ours-US-32} & 87.9\,/\,94.9\,/\,98.7 & 53.9\,/\,65.4\,/\,81.2 & \hls{33.8}\,/\,48.5\,/\,59.1 & 21.4\,/\,37.4\,/\,46.6\\

{Ours-SS-32} & \hls{88.6}\,/\,94.8\,/\,98.3 & 53.9\,/\,68.1\,/\,79.6 & 33.3\,/\,48.5\,/\,59.6 & 25.2\,/\,38.2\,/\,45.0\\

{Ours-SV-32} & 88.3\,/\,\hls{95.1}\,/\,\hls{98.9} & \hls{55.0}\,/\,69.1\,/\,\hls{84.8} & 33.3\,/\,\hls{51.0}\,/\,\hls{61.1} & \hls{28.2}\,/\,\hls{42.0}\,/\,\hls{47.3}\\ \hline

{PCA-64} & 88.7\,/\,95.6\,/\,\hlf{99.3} & 60.7\,/\,78.5\,/\,\hlf{93.2} & 35.4\,/\,56.1\,/\,68.2 & 30.5\,/\,45.0\,/\,53.4\\ 

{Ours-US-64} & 88.8\,/\,95.5\,/\,99.2 & 57.6\,/\,\hls{80.1}\,/\,92.1 & \hlf{40.9}\,/\,56.6\,/\,\hlf{71.2} & \hlf{32.1}\,/\,45.8\,/\,51.9\\ 

{Ours-SS-64} & 89.1\,/\,95.3\,/\,99.2 & 61.3\,/\,79.6\,/\,91.1 & 35.9\,/\,53.0\,/\,66.7 & 31.3\,/\,45.8\,/\,\hlf{55.7}\\ 

{Ours-SV-64} & \hlf{89.2}\,/\,\hlf{95.9}\,/\,99.2 & \hls{62.8}\,/\,79.1\,/\,92.1 & 38.9\,/\,\hlf{58.6}\,/\,70.2 & 30.5\,/\,\hlf{48.1}\,/\,54.2\\ \hline



\end{tabular}

}
\end{center}
\vspace{-3mm}
\caption{Evaluation of the localization on the Aachen and InLoc datasets with HardNet~\cite{hardnet2017} features. 
}
\vspace{-5mm}
\label{tab:loc_results:comparison_hardnet}
\end{table}

\subsection{Visual Localization}
\label{exp-loc}

In the first experiment, we evaluate how well our reduced features perform in robot visual localization tasks. We analyze different methods using two challenging localization datasets, with severe illumination changes and complex indoor scenes. We use the hierarchical localization toolbox~\cite{sarlin2019coarse} to achieve visual localization, but replace the feature extractors with generated lower-dimensional descriptors. 

For the first day-night localization challenge, we evaluate different dimensionality reduction methods on the Aachen Day-Night v1.1 dataset~\cite{9229078}. It contains $6\,697$ day-time database images from an old European town and $1\,015$ queries ($824$ taken in day and $191$ in night conditions). We use the code and evaluation protocol from~\cite{9229078} and report the percentage of day-night queries localized within a given error bound on the estimated camera position and orientation.
For the complex indoor localization challenge, we exploit the InLoc dataset~\cite{Taira2018CVPR}. It is a challenging indoor localization dataset with large differences in viewpoint and illumination between the query and database images. We also use the code and evaluation protocol from~\cite{9229078} and report the percentage of queries localized within a given error bound on the estimated camera position.

\begin{figure}[t]
\vspace{-2mm}
\centering
\def\iwidth{.49\linewidth}
\begin{minipage}{\iwidth}
    \centering
    \small{SIFT-PCA-64}
\vspace{2mm}
\end{minipage}%
\begin{minipage}{\iwidth}
    \centering
    \small{SIFT-Ours-SV-64}
\vspace{2mm}
\end{minipage}
\begin{minipage}{\iwidth}
    \centering
    \includegraphics[width=0.98\linewidth]{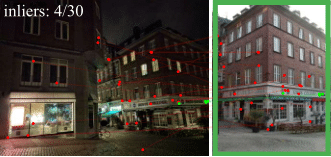}
\end{minipage}%
\begin{minipage}{\iwidth}
    \centering
    \includegraphics[width=0.98\linewidth]{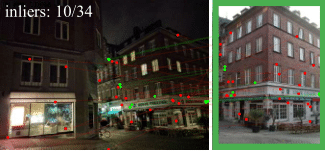}
\end{minipage}
\begin{minipage}{\iwidth}
    \centering
    \includegraphics[width=0.98\linewidth]{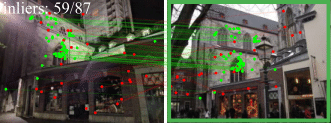}
\end{minipage}%
\begin{minipage}{\iwidth}
    \centering
    \includegraphics[width=0.98\linewidth]{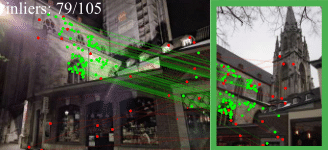}
\end{minipage}
\begin{minipage}{\iwidth}
    \centering
    \includegraphics[width=0.98\linewidth]{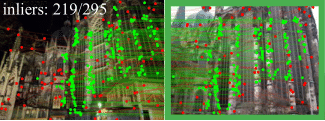}
\end{minipage}%
\begin{minipage}{\iwidth}
    \centering
    \includegraphics[width=0.98\linewidth]{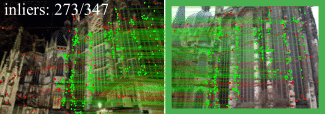}
\end{minipage}
\begin{minipage}{\iwidth}
    \centering
    \includegraphics[width=0.98\linewidth]{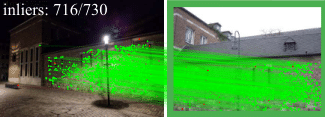}
\end{minipage}%
\begin{minipage}{\iwidth}
    \centering
    \includegraphics[width=0.98\linewidth]{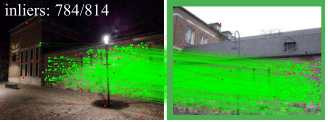}
\end{minipage}
\begin{minipage}{\iwidth}
    \centering
    \includegraphics[width=0.98\linewidth]{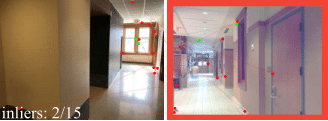}
\end{minipage}%
\begin{minipage}{\iwidth}
    \centering
    \includegraphics[width=0.98\linewidth]{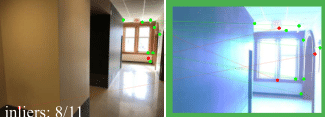}
\end{minipage}
\begin{minipage}{\iwidth}
    \centering
    \includegraphics[width=0.98\linewidth]{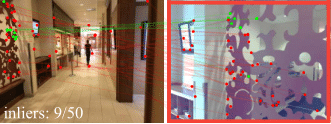}
\end{minipage}%
\begin{minipage}{\iwidth}
    \centering
    \includegraphics[width=0.98\linewidth]{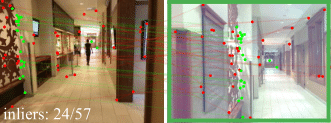}
\end{minipage}
\begin{minipage}{\iwidth}
    \centering
    \includegraphics[width=0.98\linewidth]{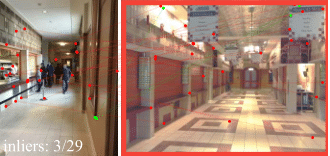}
\end{minipage}%
\begin{minipage}{\iwidth}
    \centering
    \includegraphics[width=0.98\linewidth]{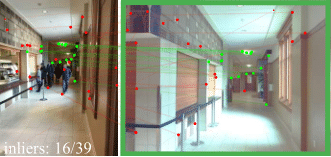}
\end{minipage}

\begin{minipage}{\iwidth}
    \centering
    \includegraphics[width=0.98\linewidth]{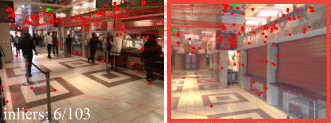}
\end{minipage}%
\begin{minipage}{\iwidth}
    \centering
    \includegraphics[width=0.98\linewidth]{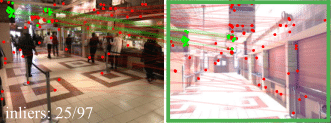}
\end{minipage}
\caption{Localization with SIFT-PCA-64 and SIFT-Ours-SV-64 on Aachen Day-Night v1.1 and InLoc. For each image pair, the left image is the query and the right image is the retrieved database image with the most inlier matches, as returned by PnP+RANSAC. Red lines represent wrong feature matches and red boxes represent wrong localization results. Green lines and boxes are correct. Best view in color and zoom in.}
\label{fig:qual:Aachen}
\vspace{-6mm}
\end{figure}

Tables~\ref{tab:loc_results:comparison} to \ref{tab:loc_results:comparison_hardnet} report the quantitative localization results and~\figref{fig:qual:Aachen} shows the qualitative results. 
For all four descriptors, our MLP-based method with auto-encoder, self-supervised, and supervised learning schemes perform better in terms of visual localization than PCA and even the original descriptors for most queries and lower dimensions. This indicates that the learned lower-dimensional descriptors are more distinctive and invariant under challenging environments, which improves the visual localization performance. 
Besides, the degree of improvement for hand-crafted descriptors is larger than the learned ones.
We will analyze and discuss this phenomenon in Sec.~\ref{sec:discuss}. 


\begin{figure}[t]
 \centering
 \includegraphics[width=0.95\linewidth]{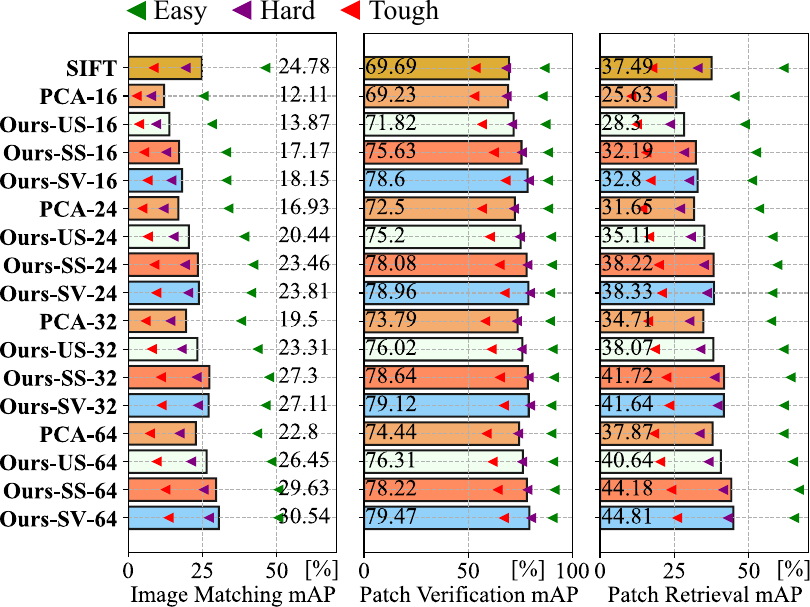}
  \caption{Verification, matching, and retrieval results of SIFT on test set `a' of HPatches dataset. None of the
MLPs is trained on HPatches. Different colors represent different difficulties and the numbers are the average mAP values.}
\label{fig:hpatches-sift}
\end{figure}

\subsection{Applications on Patch Verification, Image Matching, and Patch Retrieval}
This experiment shows more robotics applications using our dimensionality-reduced features, including the patch pair verification, image matching, and patch retrieval tasks on the HPatches dataset~\cite{hpatches2017}. There are $116$ sequences with over $1.5$ million patches in this dataset. $59$ sequences of them show significant viewpoint changes and $57$ sequences have significant illumination changes. The patches are divided into three groups: easy, hard, and tough, based on the levels of geometric noise.
Evaluation results with SIFT is shown in~\figref{fig:hpatches-sift}. 
Note that, all models of our MLP are pre-trained \emph{only} on Liberty sequence of UBC Phototour and we test them directly on HPatches dataset in a zero-shot fashion to show the good generalization of our proposed MLP-based method.

As can be seen from~\figref{fig:hpatches-sift}, our methods with supervised and self-supervised learning schemes perform better than PCA and auto-encoder for all three tasks and all lower dimensions with a large margin. 
Besides, we also observe that the $64$-dimensional descriptors generated by our method even outperform the original $128$-dimensional descriptors. Even the learned $24$-dimensional descriptors have the on par performances as the original $128$-dimensional SIFT descriptors, which shows the superiority of our proposed learned MLP-based dimensionality reduction. More explanation will be given in~\secref{sec:discuss}.

\subsection{Ablation Study}
\begin{figure}[t]
 \centering
 \centering\includegraphics[width=\linewidth]{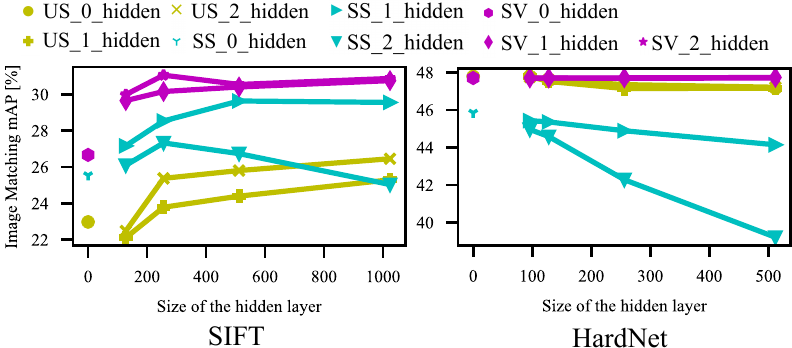}
  \vspace{-5mm}
  \caption{Impact of the number and the size of hidden layers for image matching mAP on Hatches dataset. For SIFT, the more number and size of hidden layers, the better the results are for most cases. For HardNet, the trend is opposite.}
  \vspace{-5mm}
\label{fig:ablation}
\end{figure}

In this section, we perform several experiments to provide a more in-depth analysis of how each component in MLP contributes to the final performance.
We choose different numbers of hidden layers ($0$, $1$, and $2$) with different sizes of $96$, $128$, $256$, $512$, and $1\,024$. We make evaluations on the matching task of the HPatches benchmark. We select the lower dimension of $64$ for all experiments, and~\figref{fig:ablation} shows the ablation study results with SIFT and HardNet. We can see that the more number and size of hidden layers, the better the results for the hand-crafted descriptor. However, for the learned descriptor, more hidden layers do not help. 


\subsection{Runtime and Memory Evaluation}
For each image patch, our MLP-based method adds less than $0.002\,\mathrm{ms}$ extra computational time compared to calculating the original descriptors, which can be basically ignored. Meanwhile, the lower-dimensional descriptors can save the memory $2$, $4$, $5$, and $8$ times respectively for the $64$, $32$, $24$, and $16$-dimensional descriptors, which shows the advantages of the proposed MLP-based method.

\section{Discussion}
\label{sec:discuss}

From the above experiments, we find that our learned MLP projections using unsupervised, supervised, and self-supervised methods achieve better performances than reduction using PCA. Furthermore, our features are partially even better than the original 128-dimensional ones. The degree of improvement for different descriptors is different. A key result is that the improvement for hand-crafted descriptors, like SIFT and MKD, is larger than those of learned ones, like TFeat and HardNet. Moreover, the performance with hand-crafted descriptors benefits from complex network architectures, while for learned ones, simple network architectures achieve better results.

We interpret the results as follows. 
For SIFT and MKD, the descriptor space is not optimized for the $\ell_2$ metric because of its hand-crafted design, and there are overlaps between non-matching patches. When applying PCA directly on them, matching and non-matching patches will still overlap. However, after learning a more discriminative representation using triplet loss, the projected descriptor space will be more distinctive for the lower-dimensional descriptors. This might also be why our method's performance on hand-crafted descriptors benefits from the complex network architecture - more parameters are needed to rearrange the descriptor space. This is also the reason that we do not need additional distance loss for hand-crafted ones.

For learning-based descriptors, such as HardNet and TFeat, a relatively discriminative descriptor space has already been learned using triplet loss. Similar features are close in the descriptor space, otherwise further apart. Therefore, this kind of distinctive distance information may also be preserved after the PCA projection. This might be the reason why different dimensionality reduction methods perform similarly on learning-based descriptors. However, since the descriptor space is already quite regular, it is very easy for an MLP to overfit. Thus simple architectures obtain better results for dimensionality reduction. We added an additional distance loss for HardNet to restrict the learned descriptor space and avoid overfitting. Since HardNet is trained with more advanced hard-negative mining techniques than TFeat, the performance of TFeat can still be improved significantly using our supervised MLP-based method. Due to the page limitation, we put more results and discussion in our code repository: \href{https://github.com/PRBonn/descriptor-dr}{https://github.com/PRBonn/descriptor-dr}. 

\section{Conclusion}
In this paper, we investigated an MLP-based network with unsupervised, self-supervised, and supervised learning schemes for dimensionality reduction of local feature descriptors. We thoroughly evaluate our method on four descriptors including hand-crafted and learning-based on multiple datasets with various downstream tasks. The experimental results show that our MLP-based projections work better than PCA in challenging tasks, including visual localization, patch verification, image matching, and patch retrieval for most cases. Besides, the $64$-dimensional descriptors generated from learning-based projections even outperform the original $128$-dimensional descriptors.
We also provided ablation studies and analyzed the degree of improvement for different descriptors in terms of the distribution of the descriptor space. 
Additional memory and runtime experiments show that learned lower-dimensional descriptors can be used for saving memory consumption without adding extra runtime and thus are useful for real-world robotics applications.




\bibliographystyle{plain_abbrv}
\bibliography{egbib}


\end{document}